# AI to Identify Strain-sensitive Regions of the Optic Nerve Head Linked to Functional Loss in Glaucoma

Thanadet Chuangsuwanich, PhD[1,3,8], Monisha E. Nongpiur, MD, PhD[2,3], Fabian A. Braeu, PhD[2,3,6], Tin A. Tun, MD, PhD,[2,3] Alexandre Thiery, PhD[4], Shamira Perera,[2,3] Ching Lin Ho, MD,[2,3] Martin Buist, PhD[6], George Barbastathis, PhD[6,7], Tin Aung, MD, PhD[1,2,3] and Michaël J.A. Girard, PhD[2,3,8,9,10]

[1]Yong Loo Lin School of Medicine, National University of Singapore, Singapore
[2]Duke-NUS Medical School, Singapore, Singapore.
[3]Singapore Eye Research Institute, Singapore National Eye Centre, Singapore
[4]Department of Statistics & Data Science, National University of Singapore, Singapore
[5]Department of Biomedical Engineering, National University of Singapore, Singapore
[6]Singapore-MIT Alliance for Research and Technology, Singapore
[7]Department of Mechanical Engineering, Massachusetts Institute of Technology, Cambridge, Massachusetts, USA
[8]Department of Ophthalmology, Emory University School of Medicine, Atlanta, Georgia USA
[9]Department of Biomedical Engineering, Georgia Institute of Technology/Emory University, Atlanta, GA, USA
[10]Emory Empathetic AI for Health Institute, Emory University, Atlanta, GA, USA

**Short Title:** Strain-sensitive Regions of the Optic Nerve Head

| | |
|---|---|
| **Word count:** | 3295 (manuscript text) |
| | 352 (abstract) |
| **Tables:** | 1 |
| **Figures:** | 3 |
| **Supplementary Material:** | 1 |

**Conflict of Interest:** Michael JA Girard and Alexandre Thiery are the co-founders of the AI start-up company Abyss Processing Pte Ltd

**Corresponding Authors:**

Michaël J.A. Girard, Ophthalmic Engineering & Innovation Laboratory, Emory Eye Center, Emory School of Medicine, Emory Clinic Building B, 1365B Clifton Road, NE, Atlanta GA 30322.
Email: mgirard@ophthalmic.engineering

# Abstract

**Importance:** Optic nerve head (ONH) biomechanics may drive glaucoma progression, but current assessments rely on morphology alone. Identifying strain-sensitive ONH regions could improve prediction of visual field loss and inform personalized management.

**Objective:** (1) To assess whether ONH biomechanics, quantified by tissue strain, improves prediction of three progressive visual field (VF) loss patterns in glaucoma beyond morphology alone; (2) to use explainable AI to identify strain-sensitive ONH regions contributing to these predictions.

**Design, Setting and Participants:** This is a population-based, cross-sectional study consisting of 237 glaucoma subjects. The ONH of one eye was imaged under two conditions: (1) primary gaze and (2) primary gaze with IOP elevated to ~35 mmHg via ophthalmo-dynamometry. Glaucoma experts classified the subjects into four categories based on the presence of specific visual field defects: (1) superior nasal step (N=26), (2) superior partial arcuate (N=62), (3) full superior hemifield defect (N=25), and (4) other/non-specific defects (N=124).

**Main Outcome Measures:** Classification accuracy for each VF defect type; spatial maps highlighting ONH regions most contributing to classification.

**Methods:** Automatic ONH tissue segmentation and digital volume correlation were used to compute IOP-induced neural tissue and lamina cribrosa (LC) strains. Biomechanical and structural features were input to a Geometric Deep Learning model (PointNet). Three classification tasks were performed to detect: (1) superior nasal step, (2) superior partial arcuate, (3) full superior hemifield defect. For each task, the data were split into 80% training and 20% testing sets. Area under the curve (AUC) was used to assess performance. Explainable AI techniques were employed to highlight the ONH regions most critical to each classification, providing insight into biomechanical and structural contributors to VF loss.

**Results:** Models achieved high AUCs of 0.77-0.88, showing that ONH strain improved VF loss prediction beyond morphology alone. The inferior and inferotemporal rim were identified as key strain-sensitive regions, contributing most to visual field loss prediction and showing progressive expansion with increasing disease severity.

**Conclusion and Relevance:** ONH strain enhances prediction of glaucomatous VF loss patterns. Neuroretinal rim, rather than the LC, was the most critical region contributing to model predictions, suggesting that strain at the rim could play a dominant role in the biomechanical cascade leading to axonal injury and functional loss.

# Introduction

Our recent patient studies[1-6] have demonstrated a consistent association between intraocular pressure (IOP)-induced deformations in the optic nerve head (ONH) and patterns of visual field loss, supporting a biomechanical contribution to glaucoma pathogenesis, as echoed by others.[7-12] However, while this association is increasingly accepted, direct evidence that biomechanics drives functional decline remains limited, and its predictive value for visual field progression is still uncertain.

A key challenge lies in identifying what biomechanical factors, such as stress, strain, or specific deformation modes, drive axonal injury, and where on the ONH such damaging forces are most likely to occur. Some propose the retina as the initial site, others point to the lamina cribrosa,[13] and some suggest both.[14] Clarifying the nature and location of these biomechanical triggers is essential to understanding how axonal degeneration unfolds in glaucoma and may inform more targeted approaches to patient risk assessment.

To address these questions, we strongly believe that recent AI approaches, specifically geometric deep learning (to learn from complex 3D structural and biomechanical data)[15] and explainable AI (to provide interpretability),[16] offer powerful tools to help unveil axonal injury mechanisms in glaucoma. These methods hold strong potential to identify the specific types of mechanical stimuli (whether tension, compression, or torsion) and their precise locations most likely responsible for driving visual field loss. Importantly, such techniques have already shown promise in glaucoma research by establishing critical links between ONH biomechanics and visual function.[17]

In this study, we aimed to: (1) assess whether ONH biomechanics, quantified through tissue strain, improves the prediction of distinct patterns of visual field loss that reflect progressive stages of glaucoma; and (2) use explainable AI to identify the

ONH regions most associated with these predictions, providing insights into potential biomechanical pathways of axonal injury and disease progression.

## Methods

### *Subjects Recruitment*

This study was approved by the SingHealth Centralized Institutional Review Board and adhered to the tenets of the Declaration of Helsinki. Written informed consent was obtained from each subject. This is a clinic-based cross-sectional study.

We recruited 238 glaucoma subjects from clinics at the Singapore National Eye Centre. We included subjects aged more than 50 years old, of Chinese ethnicity (predominant in Singapore), with a refractive error of ±3 diopters, and excluded subjects who underwent prior intraocular/orbital/brain surgeries, subjects with past history of strabismus, ocular trauma, ocular motor palsies, orbital/brain tumors; with clinically abnormal saccadic or pursuit eye movements; subjects with known carotid or peripheral vascular disease; or with any other abnormal ophthalmic and neurological conditions; subjects with visual field deficits related to diabetic retinopathy or any other optic neuropathies or with advanced glaucoma (mean deviation of < –25 decibels [dB]) were excluded.

Glaucoma cases were defined by the presence of glaucomatous optic nerve damage, defined as vertical cup-disc ratio (VCDR) of > 0.7 and/or neuroretinal rim narrowing with an associated visual field defect on standard automated perimetry. The latter was defined if the following were found: (1) glaucoma hemifield test outside normal limits, (2) a cluster of ≥ 3, non-edge, contiguous points on the pattern deviation plot, not crossing the horizontal meridian with a probability of < 5% being present in age-matched normals (one of which was < 1%) and (3) Pattern Standard Deviation (PSD) < 0.05; these were repeatable on two separate occasions.

Eyes with angle closure were excluded based on gonioscopy using a Sussman 4-mirror lens (Ocular Instruments, Inc., Bellevue, WA) in a dark room. The anterior chamber angle was graded according to visible anatomical structures: grade 0 (no angle structures), grade 1 (Schwalbe line), grade 2 (anterior trabecular meshwork), grade 3 (posterior trabecular meshwork or scleral spur), and grade 4 (visible ciliary body). A quadrant was considered "closed" if the posterior trabecular meshwork was not visible in the primary position without indentation (grades 0, 1, or 2), and angle closure was defined as having two or more closed quadrants.

### *Visual Field Testing*

Unreliable visual field test results with a false-positive error of more than 15% or a fixation loss of more than 33% were excluded. The thresholding algorithm (Swedish interactive testing algorithm) standard 24-2 program was selected. For every selected patient, the pattern standard deviation map from the Humphrey visual field report was used in the study. For each point in the pattern standard deviation (PSD) map, we defined a defect as a point having a p-value of less than 5%.

### *Classification of visual field defect patterns*

Our glaucoma experts (M.N. and T.A.T.) classified visual field defect (VFD) patterns based on each patient's visual field report. The classification system used was consistent with our previous work,[18] which was derived from the Ocular Hypertension Treatment Study classification system.[19]

Subjects were categorized into the following groups based on their initial visual field defect (VFD), excluding the first visual field test: (1) Presence of a superior nasal step; (2) Presence of a superior arcuate defect; (3) Presence of a full superior hemifield defect; (4) Other types of VFDs, including inferior or nonspecific defects.

We focused on superior VFDs, as they were the most prevalent in our dataset and are commonly observed in early glaucomatous damage.[20] These categories represent a spectrum of superior visual field loss, ranging from mild (nasal step) to more advanced damage (full hemifield defect).

## *OCT Imaging and Biomechanical Testing*

We selected one eye at random from each patient, and we imaged the ONH with spectral-domain OCT (Spectralis; Heidelberg Engineering GmbH). The imaging protocol was similar to that from our previous work.[3] In brief, we conducted a raster scan of the ONH (covering a rectangular region of 150 × 100 centered at the ONH) comprising 97 serial B-scans, with each B-scan comprising 384 A-scans. The average distance between B-scans was 35.1 µm, and the axial and lateral (distance between a-scans) resolution on average were 3.87 µm and 11.5 µm, respectively. Each eye was scanned twice under 2 conditions: OCT at baseline and under acute IOP elevation. Each patient was administered 1.0% tropicamide to dilate the pupils before imaging.

For each eye in baseline position, we applied a constant force of 0.65 N to the temporal side of the lower eyelid using an ophthalmodynamometer (ODM), as per a well-established protocol.[1,3,21] This force raised IOP to approximately 35 mmHg and was maintained constant. IOP then was reassessed with a TonoPen (Reichert Instruments GmbH), and the ONH was imaged with OCT immediately (within 30 seconds) after the IOP was measured.

## AI-based Segmentation of ONH Tissues and Representation of the ONH Structure as 3D Point Cloud

We automatically segmented all baseline OCT volume scans of the ONH using REFLECTIVITY (Reflectivity, Abyss Processing Pte Ltd, Singapore). More specifically, the following ONH tissue groups were automatically labelled (**Figure 1a**): (1) the RNFL and the prelamina tissue; (2) the ganglion cell inner plexiform layer; (3) all other retinal layers; (4) the retinal pigment epithelium with Bruch's membrane (BM) and the BM opening (BMO) points; and (5) the OCT-visible part of the lamina cribrosa (LC). In almost all OCT volume scans, the posterior boundaries of the LC were not visible and could not be segmented.

Each ONH was then represented as a 3D point cloud derived from the segmentation of OCT images and served as input to our geometric deep learning algorithm, following an approach established in our previous work. All ONH point clouds were rigidly aligned with respect to Bruch's membrane opening (BMO) center and plane to ensure consistent anatomical orientation across subjects. Full methodological details are provided in the Appendix.[22-24]

## Deformation Tracking and IOP-induced Effective Strain Derivation

To extract the 3D deformation of the ONHs under an acute elevation in IOP, we used a commercial digital volume correlation (DVC) module (Amira (version: 2020.3), Thermo Fisher Scientific, USA). Details about the preprocessing of the volumes and DVC algorithm can be found in our previous work.[3] From the deformation mapping, we computed the effective strain within the ONH, a measure of local 3D deformation that accounts for both compressive and tensile components – higher values indicate greater mechanical deformations in either direction. We then assigned effective strain values to each point in the ONH point cloud using K-nearest neighbor interpolation (K = 5) (**Figure 1b**).

## *Classification of Visual Field Defects Using Geometric Deep Leaning*

In this study, we used PointNet to take each ONH point cloud (example in **Figure 1b**) as input to predict the corresponding visual field defect map (52 points) for that specific eye.[23]

We trained three separate PointNet models, each targeting a distinct classification task:

(1) Detecting the presence of a superior nasal defect (Presence: N = 26; Absence: N = 211);

(2) Detecting the presence of a superior arcuate defect (Presence: N = 62; Absence: N = 175);

(3) Detecting the presence of a full superior hemifield defect (Presence: N = 25; Absence: N = 212).

Each model was trained to classify the presence or absence of a specific defect pattern based on 3D ONH morphology and biomechanical strain. The network was trained using binary cross-entropy loss, with the dataset split into 80% training and 20% testing. Due to the limited number of subjects with the specified defect for each task, no validation set was used in this study. Model hyperparameters were tuned based on our previous work,[17] and the final iteration of the trained weights was used for evaluation on the test set. Model performance was assessed by reporting the area under the curve (AUC) for each task's test set.

During the training process, we augmented the point clouds in each sample using various techniques: random cropping, random rotations, and random sampling of points (selecting a subset of 3,000 points from each point cloud).[25] This augmentation enhances the model's generalizability to unseen datasets.

## *Explainable AI to Identify Strain-sensitive Regions of the ONH Linked to Functional Loss*

To identify the ONH regions most critical for prediction, we generated patient-averaged saliency maps by computing the gradient of the model output with respect to each input point. Gradient magnitudes across spatial, structural, and strain features were averaged to reflect each point's importance. Because saliency maps are represented in 3D, we simplified their visualization. Saliency values (gradients) were sum-projected onto an en-face plane (2 mm × 2 mm grid centered on BMO, with 40 × 40 cells) and also visualized along inferior-superior and nasal-temporal ONH cross-sections. Note that all ONHs were scaled relative to the BMO radius to ensure consistent anatomical representation across subjects, and all saliency maps were normalized to a scale between 0 and 1. Saliency maps were generated for each of the three classification tasks: detection of nasal steps, arcuate defects, and hemifield defects.

Finally, from the saliency maps, we defined regions of high gradient as strain-sensitive regions – area where local strain values most strongly contribute to the prediction of characteristic visual field loss patterns in glaucoma.

## *Importance of Biomechanical Strain in Predicting Visual Field Loss*

To assess the contribution of biomechanical strain features to model performance, we conducted a sensitivity analysis focused on the classification of superior arcuate visual field defects. This pattern was selected because it represents a well-characterized and relatively common form of early glaucomatous damage,[20,26] and, critically, was the only superior defect subtype in our dataset with a sufficiently balanced sample size (Presence: N = 62; Absence: N = 175). In contrast, the superior nasal step (N = 26) and full superior hemifield defect (N = 25) groups had markedly fewer cases.

We trained the model with and without strain features and evaluated performance across five independent train-test splits. Results were reported as the area under the receiver operating characteristic curve (AUC), expressed as mean ± standard deviation.

## Results

### Subjects' demographics

A total of 237 Chinese subjects with glaucoma were recruited for this study, with a mean age of 69 ± 5 years; 128 of them were female. The cohort spanned a broad spectrum of disease severity, with Mean Deviation (MD) values ranging from -1.8 dB (mild) to -25.2 dB (severe), and an average MD of -7.25 ± 5.05 dB. Among these subjects, 26 presented with a nasal step defect, 62 had a superior partial arcuate defect, 25 exhibited a full superior hemifield defect, and the remaining 124 had other or non-specific visual field abnormalities. A detailed summary of demographic and clinical characteristics stratified by visual field defect subtype is provided in **Table 1**.

### Model Performance Across Visual Field Defect Classifications

Overall, our geometric deep learning model demonstrated strong performance across all classification tasks, achieving peak AUC values of 0.77 for nasal step detection, 0.88 for superior partial arcuate defects, and 0.87 for superior hemifield defects.

## Incorporation of Effective Strains Improves Prediction of Superior Arcuate Pattern

We found that the model incorporating biomechanical strain significantly outperformed the model without strain in predicting superior arcuate defects (AUC: $0.87 \pm 0.02$ vs. $0.83 \pm 0.02$, $p < 0.05$). These results suggest that local biomechanical strain is a meaningful predictor of localized visual field loss and may play an important role in the pathophysiology of region-specific glaucomatous damage.

## Explainable AI Reveals an Arching Pattern in the Infero-Temporal Neuroretinal Rim, Increasing with Glaucoma Severity

Across all models, critical regions from the saliency maps (also defined as strain-sensitive regions) consistently appeared in the inferior and infero-temporal neuroretinal rim (Figure 2). In the en-face view, a spreading pattern was observed, extending from the inferior to the temporal region (Figure 2d-e), forming an arc that increased in length as the defect progressed from a nasal step to a arcuate pattern and, ultimately, to full hemifield loss (all superior). As the arc expanded, it covered an increasingly larger proportion of axons entering the disc. Additionally, a less prominent critical region was observed in the superior neuroretinal rim, which may reflect inferior visual field defects in some patients.

## Discussion

In this study, we demonstrated that incorporating ONH tissue strain into AI models significantly improved the prediction of visual field loss patterns in glaucoma beyond what was possible using structural features alone. Using geometric deep learning on 3D point clouds of the ONH and acute IOP-induced strain measurements, we achieved high classification performance across three visual field defect types relevant to glaucoma progression. Importantly, explainable AI consistently identified the inferior and infero-temporal neuroretinal rim as the most critical region contributing to these predictions. These findings raise the possibility that local IOP-induced strain in the neuroretinal rim is associated with axonal injury and may help explain patterns of region-specific visual field loss in glaucoma.

In this work, we found that incorporating biomechanical strain information significantly improved the prediction of a localized visual field defect. Unlike our earlier work, which demonstrated that strain enhances prediction of full-field visual sensitivity maps,[17] the current approach targeted a specific defect type (i.e., arcuate) commonly observed in glaucoma.[27] Specifically, we were able to reveal a more spatially resolved link between IOP-induced mechanical deformation and site-specific visual loss. This localized relationship is biologically plausible, as strain has been suggested to (1) cause direct mechanical injury to retinal ganglion cells (RGCs),[28,29] (2) disrupt microcapillary blood flow in the lamina cribrosa, choroid, and retina,[30,31] and (3) impair axoplasmic transport.[32] These deformation-driven pathways compromise axonal integrity and contribute to vision loss in glaucoma. Our findings suggest that local biomechanical strain is not only a global predictor of visual function but also provides valuable insight into the region-specific vulnerability of the ONH.

With explainable AI, we found that strain-sensitive regions of the ONHs increased in size with glaucoma severity and were consistently concentrated around the inferior and infero-temporal neuroretinal rim. Specifically, for eyes with superior nasal defects, our AI models consistently identified the inferior neuroretinal rim as the most critical region. As visual field loss progressed into superior arcuate and then superior full hemifield defects, these critical regions expanded to include the inferotemporal and inferonasal rims. One way to interpret these findings is that, as glaucoma progresses, strain affecting an increasing proportion of axons entering the optic disc becomes predictive of expanding visual field defects, which could be consistent with a developing pattern of axonal injury. Confirming this hypothesis will require longitudinal monitoring of patients undergoing biomechanical testing, which is currently underway in our laboratory.

Importantly, these transitions were entirely data-driven and emerged naturally from the model, without reliance on any predefined anatomical maps. Yet, the spatial relevance patterns identified by our AI models align well with established structure-function relationships. For instance, they correspond with the Garway-Heath map,[33] which links superior nasal field loss to damage in the inferior rim and associates more extensive visual field defects with broader regions of the optic disc. Unlike the Garway-Heath framework, however, our method does not rely on pre-defined sectors. Instead, it uses geometric deep learning and explainable AI to generate patient-specific saliency maps that capture biomechanics-function associations – offering a potential key to uncovering the mechanisms of axonal injury.

Another notable observation from our saliency map (Figure 2) is the lower importance assigned to the LC region compared to the prelaminar tissue. This is intriguing, as the LC undergoes significant remodeling changes in glaucoma, such as excavation and increased curvature.[34-36] Additionally, the LC is a key region where biomechanical strain accumulates,[37] a factor that has been linked to glaucoma progression in both our study and previous research.[3,38] Given these, we would expect the LC to have greater importance, particularly in cases of more severe glaucomatous damage, such as hemifield defects. This discrepancy may arise from several factors: limitations in how the model interprets the lamina cribrosa's role, constraints in feature representation learning within PointNet,[39] insufficient resolution of LC pores using current clinical OCT technology,[40] or the small number of positive samples for certain defect types in our dataset. Alternatively, it may reflect a true biological sensitivity, as axons in the neuroretinal rim must undergo a sharp 90-degree turn – a potential point of mechanical vulnerability.[41] Advances in OCT imaging may help clarify these possibilities.

In this study, several limitations warrant further discussion. First, our AI models only considered effective strain as the biomechanical input. While effective strain captures the magnitude of local deformation, it does not distinguish between distinct mechanical modes, such as stretching, compression, or shear. As a result, the specific nature of the biomechanical insult remains unclear, ultimately limiting our ability to identify the precise mechanisms that could underly axonal injury. We are currently developing models that incorporate full strain and stress fields to provide a more comprehensive understanding of the mechanical forces that could contribute to visual field loss.

Second, our geometric deep learning model (PointNet) has limited ability to disentangle the regional contributions of individual features, such as strain versus tissue thickness, within the saliency map. This limitation arises because PointNet combines features in its early layers, making it difficult to isolate their specific effects. We are currently developing an AI method to separately analyze the influence of strain, morphological parameters, and vascular features to gain clearer insight into their respective contributions and interactions.

Third, our sample size is limited and the dataset is unbalanced, particularly for subjects with specific defect patterns (e.g., only 25 individuals with superior hemifield defects). This limitation may affect the generalizability of our findings and underscores the need for a larger, more diverse dataset in future studies. Moreover, the visual field defect types we examined, superior nasal step, superior arcuate, and superior hemifield loss, do not preclude the presence of additional defects in the inferior region, which may explain the less prominent critical regions observed in the superior neuroretinal rim (Figure 2d–f). In future work, we aim to recruit a larger cohort and include a broader spectrum of visual field loss patterns to enable a more comprehensive evaluation of the saliency maps.

In conclusion, our findings demonstrate that ONH tissue strain enhances the prediction of distinct visual field loss patterns typically observed in glaucoma. Explainable AI revealed that the neuroretinal rim, rather than the LC, was the most critical region contributing to model predictions, suggesting that strain at the rim could play a dominant role in the biomechanical cascade leading to axonal injury and functional loss.

## Acknowledgments

We acknowledge funding from **(1)** the donors of the National Glaucoma Research, a program of the BrightFocus Foundation, for support of this research (G2021010S [MJAG]); **(2)** the "Retinal Analytics through Machine learning aiding Physics (RAMP)" project that is supported by the National Research Foundation, Prime Minister's Office, Singapore under its IntraCreate Thematic Grant "Intersection Of Engineering And Health" - NRF2019-THE002-0006 awarded to the Singapore MIT Alliance for Research and Technology (SMART) Centre [MJAG/AT/GB], **(3)** the

NMRC-LCG grant 'TAckling & Reducing Glaucoma Blindness with Emerging Technologies (TARGET)', award ID: MOH-OFLCG21jun-0003 [MJAG], **(4)** the Emory Eye Center [Start-up funds, MJAG], **(5)** Support from a Challenge Grant from Research to Prevent Blindness, Inc. to the Department of Ophthalmology at Emory University; and **(6)** Support from the NIH grant P30EY06360 to the Atlanta Vision Community.

| Characteristic (mean ± standard deviation) | Subjects with superior nasal defect | Subjects with superior arcuate defect | Subjects with superior hemifield defect | Subjects with non-specified or other defects |
|---|---|---|---|---|
| Age (year) | 69 ± 6 | 69 ± 6 | 70 ± 4 | 70 ± 5 |
| Sex, female (%) | 79% | 54% | 52% | 64% |
| Axial Length (mm) | 24.0 ± 1.0 | 24.0 ± 1.0 | 23.8 ± 0.8 | 23.7 ± 1.0 |
| Visual field, MD (dB) | -7.5 ± 4.6 | -8.0 ± 4.4 | -14.0 ± 3.6 | -9.4 ± 5.3 |
| Pattern standard deviation (dB) | 7.9 ± 4.9 | 9.0 ± 4.0 | 13.0 ± 2.7 | 9.7 ± 5.5 |
| Baseline IOP (mmHg) | 15.8 ± 3.4 | 15.8 ± 3.0 | 14.7 ± 2.7 | 15.0 ± 3.7 |
| IOP (mmHg) with indentation^ | 35.9 ± 4.4 | 33.8 ± 6.4 | 35.0 ± 5.5 | 34.8 ± 5.9 |

**Table 1**: Subjects' demographics.

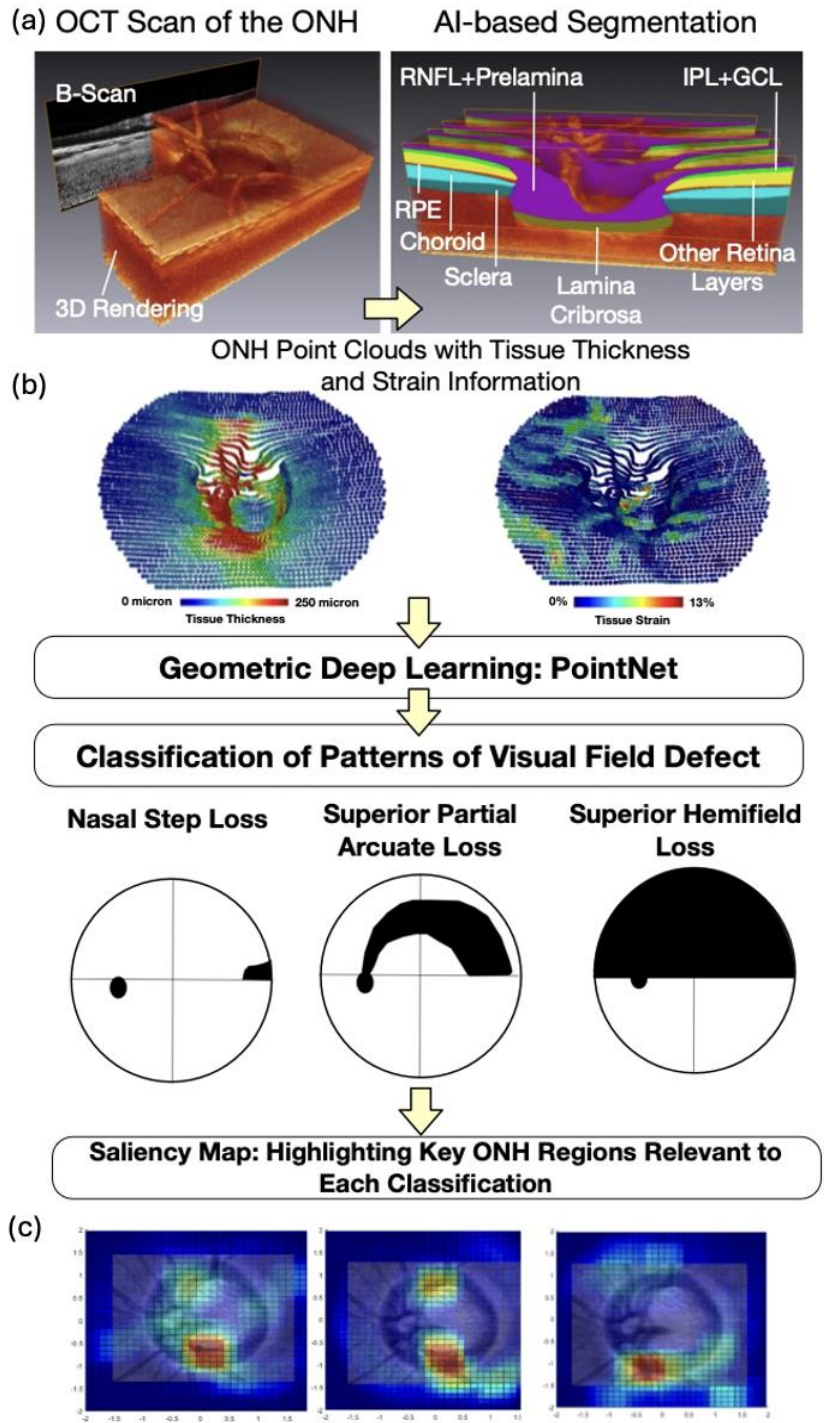

**Figure 1(a)** Segmentation of ONH neural tissues consisting of the retinal nerve fiber layer (RNFL) and the prelamina tissue (PLT), ganglion cell inner plexiform layer (GCL+IPL), all other retinal layers (ORL), the retinal pigment epithelium (RPE) and the lamina cribrosa (LC). **(b)** An illustration of ONH point cloud containing both the tissue thickness and the effective strain as an input to the PointNet. The output is the binary classification of three separate classification tasks. **(c)** An example of a saliency map representing the importance of each region to each classification task.

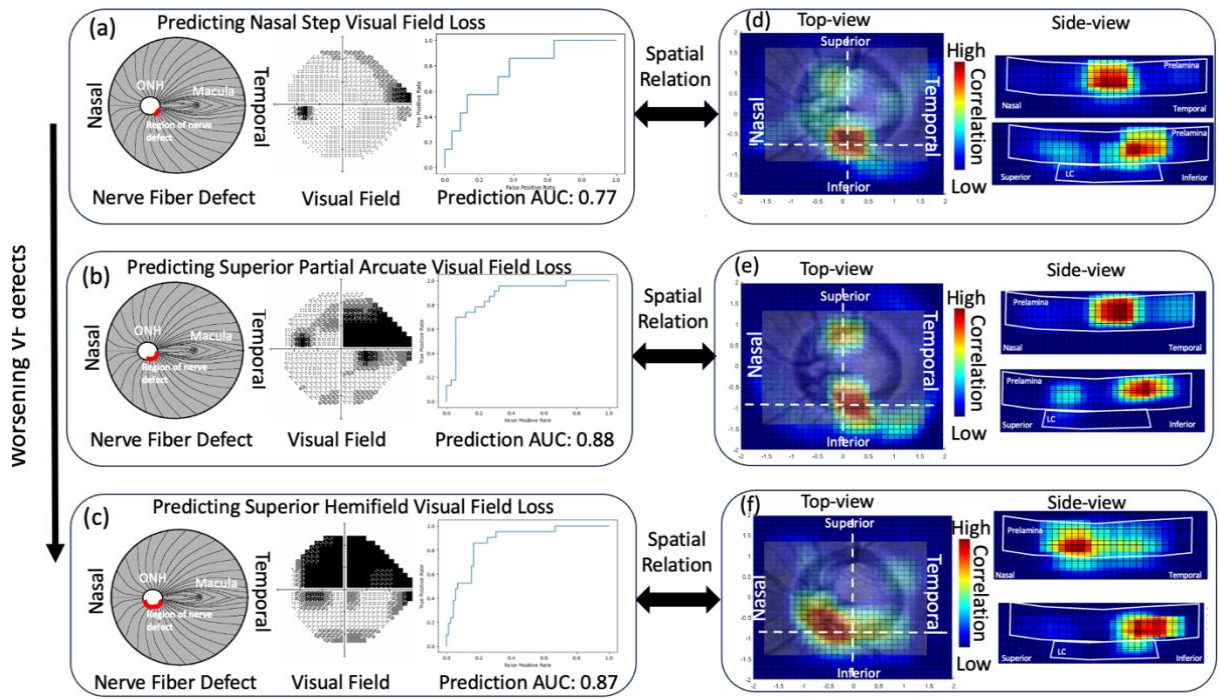

**Figure 2.** Comparison of Visual Field Defect Patterns by Severity: (a) Nasal step, (b) superior arcuate, and (c) superior hemifield loss are illustrated with corresponding visual field defect samples from our population. Red regions in the nerve fiber defect maps denote areas of nerve loss, aligned with established structure-function relationships such as the Garway-Heath map. (d-f) Saliency maps highlight an arching pattern, where the arc length increases with defect severity. Red regions indicate areas of high correlation with each defect pattern. To provide anatomical context, a sample ONH is overlaid in the background, referencing the neuroretinal rim. Additional cross-sectional views are provided: one along the superior-inferior axis through the ONH's midsection, and another along the nasal-temporal direction at the neuroretinal rim.